\documentclass[letterpaper]{article} % DO NOT CHANGE THIS
\usepackage{aaai24}  % DO NOT CHANGE THIS
\usepackage{times}  % DO NOT CHANGE THIS
\usepackage{helvet}  % DO NOT CHANGE THIS
\usepackage{courier}  % DO NOT CHANGE THIS
\usepackage[hyphens]{url}  % DO NOT CHANGE THIS
\usepackage{graphicx} % DO NOT CHANGE THIS
\usepackage{natbib}  % DO NOT CHANGE THIS AND DO NOT ADD ANY OPTIONS TO IT
\usepackage{caption} % DO NOT CHANGE THIS AND DO NOT ADD ANY OPTIONS TO IT
\usepackage{algorithm}
\usepackage[dvipsnames, svgnames, x11names]{xcolor}
\usepackage{colortbl}
\usepackage{multirow}
\usepackage{subfigure}
\usepackage{newfloat}
\usepackage{listings}
\usepackage{times}
\usepackage{epsfig}
\usepackage{graphicx}
\usepackage{amsmath}
\usepackage{amssymb}
\usepackage{color}
\usepackage[mathscr]{eucal}
\usepackage[dvipsnames, svgnames, x11names]{xcolor}
\usepackage{colortbl}
\usepackage{arydshln}
\floatstyle{ruled}
\newfloat{listing}{tb}{lst}{}
\floatname{listing}{Listing}

\title{Non-Exemplar Online Class-incremental Continual Learning via \leavevmode\\ Dual-prototype Self-augment and Refinement \leavevmode\\ ——Appendix——}
\author{
    Fushuo Huo\textsuperscript{\rm 1}, Wenchao Xu\textsuperscript{\rm 1}, Jingcai Guo\textsuperscript{\rm 1, \rm 2}, Haozhao Wang\textsuperscript{\rm 3}\footnote{Corresponding author}, Yunfeng Fan\textsuperscript{\rm 1}
}
\affiliations{
    \textsuperscript{\rm 1}Department of Computing, The Hong Kong Polytechnic University, China \leavevmode\\
    \textsuperscript{\rm 2}The Hong Kong Polytechnic University Shenzhen Research Institute, Shenzhen, China \leavevmode\\
    \textsuperscript{\rm 3}School of Computer Science and Technology, Huazhong University of Science and Technology, China \leavevmode\\
}

\usepackage{bibentry}
\begin{document}

\maketitle

\section{Overview}
The appendix presents more experimental settings, results, and analyses as follows:

\textbf{Appendix A} More implementation details

\textbf{Appendix B} More results on different dataset partitions.

\textbf{Appendix C} Results and analysis on different base session training strategies.

\textbf{Appendix D} Hyper parameter analysis

\textbf{Appendix E} Computation overhead analysis

% \\ \hspace*{\fill} \\
\textbf{\large{Appendix A:  More Implementation Details}}
% \\ \hspace*{\fill} \\

\textbf{Dataset Overview.} We conduct experiments on three widely used datasets, including CORE-50 \cite{core50}, CIFAR 100 \cite{cifar100}, and Mini-ImageNet \cite{miniimagenet}. Here we give brief introductions. CORE-50 \cite{core50} is a benchmark designed for class incremental learning with 50 classes. Each class has
around 2,398 training images and 900 testing images, with the size of $3 \times 128 \times 128$. CIFAR 100 \cite{cifar100} contains 60000 images of $32 \times 32$ size from 100 classes, and each class includes 500 training images and 100 test images. Mini-ImageNet \cite{miniimagenet} contains 100 classes and is divided into 10 sub-datasets for 10 disjoint tasks, and each task contains 10 classes. Each task comprises 5,000 training images and 1,000 testing images, all with the size of $3 \times 84 \times 84$.

\noindent\textbf{Training Details.} \textbf{For OCL methods}, we employ the same dataset partitions and training protocols of NO-CL, i.e., pre-training on the base classes and then online class-incremental learning \emph{with example buffers}. Other hyperparameters are adopted as default. The example buffers are restored and retrieved during the whole training procedure with the default updating pipeline. MIR \cite{mir}, GD \cite{gdumb}, ASER \cite{sv}, SCR \cite{scr}, and DVC \cite{dvc} are based on the OCL codebase \footnote{https://github.com/RaptorMai/online-continual-learning}. Other methods are implemented with the public released codes. \textbf{For FS-CL methods}, FACT \cite{fs-cl-fact} and ALICE \cite{fs-cl-open}, we also adopt the same protocols as NO-CL. The prototypes of novel classes are computed by all data samples rather than few-shot samples. During the inference phase, FACT and ALICE directly infer incremental data samples via computed prototypes without finetuning the network. All methods are implemented with the public released codes. \textbf{For NE-CL methods} \cite{ne-cl-pa, ne-cl-ss}, as \cite{ne-cl-ss} does not provide training scripts, we adopt the three-party codes \cite{pycil}\footnote{https://github.com/G-U-N/PyCIL} on CIFAR100 dataset. For \cite{ne-cl-pa}, we employ the same dataset partitions and training and testing protocols of NO-CL. Note that all methods are employed the \emph{same} reduced ResNet-18 as the feature extractor for fair comparisons. All experiments are conducted with NVIDIA RTX3090 GPU on CUDA 11.4 using PyTorch framework.

\noindent\textbf{Base Session Training Details.} Here, we give the details of our base session training strategy. We employ base training functions on the outputs of the feature extractor and projection module to obtain vanilla and high-dimensional prototypes for sequentially online sessions: $L^{base}=L^{base}_{vp} + L^{base}_{hp}$. $L^{base}_{vp}=Loss(Proj_{vp}(\theta_{1}(x)), y)$ and $L^{base}_{vp}=Loss(Proj_{hp}(\theta{_2}(\theta_{1}(x))), y)$. $x$, $y$, $\theta_{1}$, and $\theta_{2}$ denote input samples, labels, feature extractor, and projection module. $Proj_{vp/hp}$ are linear layers to align vanilla- and high-dimensional prototypes for loss calculations. For cross-entropy (CE) loss functions, $Proj_{vp/hp}$ are one-layer MLP with the output dimension of base class. For supervised contrastive (SC) loss \cite{scl}, we follow SCR \cite{scr} and adopt the \emph{same} hyper-parameters of SCR. $Proj_{vp/hp}$ are two-layer MLP with the dimension of 160 and 128 and the temperature is set to 0.1.
% Concretely, SCL is formulated as:
% \begin{eqnarray}
% L^{scl}=\sum_{i\in I}{\frac{-1}{\left| P\left( i \right) \right|}}\sum_{p\in P\left( i \right)}{\log \frac{\exp \left( z_i\times z_p/\tau \right)}{\sum_{a\in A\left( i \right)}{\exp \left( z_i\times z_a/\tau \right)}}}
% \end{eqnarray}
% where $I$ is the set of indices of Batch ($B_{I}=\{x_{k}, y_{k}, Aug(x_{k}), y_{k}\}{_{k=1...b}}$), which consists of original batch and augmented ($Aug$) view with $2b$ samples. $A(i)=I\backslash \{ i\} $ means the set of indices of all samples in $B_{I}$ except for sample $i$.  $z_{i}=Proj(\rm{HD}$$_{ft})$.  $\rm{HD}$$_{ft}$ is the output features from hyperembedding projection and  $Proj$ is a multi-layer perceptron 
% % of size 2048 and output vector of 1024.
% with a single hidden layer. $P(i) \equiv \{ p \in A(i):{y_p} = {y_i}\} $ is the set of indices of all positives in the augmented batch distinct from $i$, where $|P(i)|$ is its cardinality. $\tau \in \mathcal{R}^+$ is an temperature parameter. 

% \\ \hspace*{\fill} \\
\textbf{\large{Appendix B: Results on Different Dataset Partitions}}
% \\ \hspace*{\fill} \\

Due to space constraints in the main paper, in this subsection, we report the additional results of different dataset partitions. Concretely, as shown in Tables \ref{40-6-10} and \ref{80-2-10}, we conduct experiments on the configuration of $40\%  + 6\%  \times 10$ and $80\%  + 2\%  \times 10$, where $40\%$ and $60\%$ classes are selected as the base classes, and the rest classes are continually fed to the network in 10 sessions. Moreover, more incremental sessions (i.e., 20 sessions) with the configuration of $60\%  + 2\%  \times 20$ are also compared in Table \ref{60-2-20}. The five representative state-of-the-art methods are compared with the same training and inference protocols as Non-exemplar Online Class-incremental continual Learning (NO-CL), including Online Class-incremental continual Learning (OCL) methods (i.e., SCR \cite{scr}, OCM \cite{ocm}, and DVC\cite{dvc}), Non-Exemplar Class-incremental continual Learning (NE-CL) method (i.e., PASS \cite{ne-cl-pa}), and Few-Shot Class-incremental Learning (FS-CL) method (i.e., ALICE \cite{fs-cl-open}). As we can see from Tables \ref{40-6-10} and \ref{80-2-10}, fewer base classes result in poor performance, both in base and novel classes, because the network lacks enough pre-trained information to generalize to novel classes. Meanwhile, as our method depends on the inner-prototype computed by the pre-trained backbone, the well-trained backbone benefits us much for the prototype refinement. Notably, even with $40\%$ base classes, our method also achieves the best performance in Acc and HM metrics, which validates the robustness of prototype refinement strategies. Moreover, as for more incremental sessions in Table \ref{60-2-20}, the performance only drops slightly. Overall, experiments on different dataset partitions validate the effectiveness and robustness of our method.

\begin{table*}[]\centering
\scriptsize
\renewcommand\arraystretch{1.1}
\begin{tabular}{l|clcl|clclclcl}
\hline
\textbf{Methods} & \multicolumn{4}{c|}{\textbf{CORE-50}}                                                 & \multicolumn{4}{c}{\textbf{CIFAR100}}                                                 & \multicolumn{4}{c}{\textbf{Mini-ImageNet}}                                           \\ \hline
\textbf{Metrics} & \multicolumn{4}{c|}{\textbf{Acc}(base/novel)$|$\textbf{HM}}                                               & \multicolumn{4}{c|}{\textbf{Acc}(base/novel)$|$\textbf{HM}}                                               & \multicolumn{4}{c}{\textbf{Acc}(base/novel)$|$\textbf{HM}}                                               \\ \hline
ALICE            & \multicolumn{4}{c|}{35.0(49.3/25.5)$|$33.6}                                             & \multicolumn{4}{c|}{37.1(59.2/22.3)$|$32.4}                                             & \multicolumn{4}{c}{36.5(58.6/21.8)$|$31.8}                                             \\ \hline
PASS             & \multicolumn{4}{c|}{25.2(62.8/0.2)$|$0.4}                                                   & \multicolumn{4}{c|}{26.4(65.1/0.6)$|$1.2}                                               & \multicolumn{4}{c}{25.8(63.4/0.8)$|$1.6}                                                 \\ \hline
\textbf{MS}      & \multicolumn{2}{c|}{1000}                 & \multicolumn{2}{c|}{2000}                 & \multicolumn{2}{c|}{1000}                 & \multicolumn{2}{c|}{2000}                 & \multicolumn{2}{c|}{1000}                 & \multicolumn{2}{c}{2000}                 \\ \hdashline
SCR              & \multicolumn{2}{c|}{37.3(36.9/37.5)$|$37.2} & \multicolumn{2}{c|}{38.7(38.9/38.6)$|$38.7} & \multicolumn{2}{c|}{32.7(34.2/31.7)$|$32.9} & \multicolumn{2}{c|}{34.9(36.9/33.6)$|$35.2} & \multicolumn{2}{c|}{29.8(29.1/30.3)$|$29.7} & \multicolumn{2}{c}{36.3(39.2/34.4)$|$36.6} \\
SCR$_{ft}$            & \multicolumn{2}{c|}{34.2(45.2/26.8)$|$33.6} & \multicolumn{2}{c|}{37.7(49.8/29.6)$|$37.1} & \multicolumn{2}{c|}{30.9(45.2/21.3)$|$21.3} & \multicolumn{2}{c|}{36.4(50.8/26.8)$|$35.1} & \multicolumn{2}{c|}{32.9(43.2/26.1)$|$32.5} & \multicolumn{2}{c}{36.8(50.8/27.5)$|$35.7} \\
OCM              & \multicolumn{2}{c|}{37.6(37.5/37.6)$|$37.5} & \multicolumn{2}{c|}{39.5(40.5/38.9)$|$39.7} & \multicolumn{2}{c|}{31.6(32.9/30.8)$|$31.8} & \multicolumn{2}{c|}{37.1(36.1/37.8)$|$36.9} & \multicolumn{2}{c|}{31.3(30.8/31.6)$|$31.2} & \multicolumn{2}{c}{31.7(32.9/30.9)$|$31.9} \\
OCM$_{ft}$            & \multicolumn{2}{c|}{35.8(43.1/30.9)$|$36.0} & \multicolumn{2}{c|}{36.7(46.1/30.6)$|$36.7} & \multicolumn{2}{c|}{32.4(48.2/21.8)$|$30.1} & \multicolumn{2}{c|}{35.9(50.6/26.1)$|$36.8} & \multicolumn{2}{c|}{30.2(38.1/24.9)$|$31.4} & \multicolumn{2}{c}{37.7(51.2/28.7)$|$36.8} \\
DVC              & \multicolumn{2}{c|}{37.5(36.9/37.9)$|$37.3} & \multicolumn{2}{c|}{38.7(39.8/38.0)$|$38.9} & \multicolumn{2}{c|}{29.9(30.0/29.8)$|$29.9} & \multicolumn{2}{c|}{37.2(34.6/39.0)$|$36.6} & \multicolumn{2}{c|}{29.7(29.9/29.6)$|$29.7} & \multicolumn{2}{c}{33.3(35.6/31.8)$|$33.6} \\
DVC$_{ft}$            & \multicolumn{2}{c|}{36.7(45.8/30.6)$|$36.7} & \multicolumn{2}{c|}{37.9(45.1/33.1)$|$38.2} & \multicolumn{2}{c|}{32.3(43.6/24.7)$|$31.5} & \multicolumn{2}{c|}{36.5(46.5/29.8)$|$36.3} & \multicolumn{2}{c|}{31.4(39.4/26.1)$|$31.3} & \multicolumn{2}{c}{32.6(38.5/28.6)$|$32.8} \\ \hline
\rowcolor{WhiteSmoke}\textbf{Ours}    & \multicolumn{4}{c|}{\textbf{45.5}(44.2/46.4)$|$\textbf{45.3}}                                       & \multicolumn{4}{c|}{\textbf{38.6}(43.8/35.2)$|$\textbf{39.0}}                                    & \multicolumn{4}{c}{\textbf{40.4}(55.4/30.4)$|$\textbf{39.3}}         \\ \hline
\end{tabular}
\caption{\textbf{The quantitative analysis of dataset partition $\textbf{40\%}+\textbf{6\%}\times\textbf{10}$.} Class-wise accuracy ($\textbf{Acc}$) by end of the training in terms of all classes, base classes, and novel classes and Harmonic accuracy (\textbf{HM}) are illustrated. \textbf{MS} and $_{ft}$ mean the example memory size and finetuning versions. The best results are marked in \textbf{bold}.}
\label{40-6-10}
\end{table*}

\begin{table*}[]\centering
\scriptsize
\renewcommand\arraystretch{1.1}
\begin{tabular}{l|clcl|clclclcl}
\hline
\textbf{Methods} & \multicolumn{4}{c|}{\textbf{CORE-50}}                                                  & \multicolumn{4}{c}{\textbf{CIFAR100}}                                                 & \multicolumn{4}{c}{\textbf{Mini-ImageNet}}                                           \\ \hline
\textbf{Metrics} & \multicolumn{4}{c|}{\textbf{Acc}(base/novel)$|$\textbf{HM}}                                                & \multicolumn{4}{c|}{\textbf{Acc}(base/novel)$|$\textbf{HM}}                                               & \multicolumn{4}{c}{\textbf{Acc}(base/novel)$|$\textbf{HM}}                                               \\ \hline
ALICE            & \multicolumn{4}{c|}{44.7(48.2/30.6)$|$37.4}                                              & \multicolumn{4}{c|}{50.5(57.2/23.8)$|$33.6}                                             & \multicolumn{4}{c}{50.3(56.4/26.1)$|$35.7}                                             \\ \hline
PASS             & \multicolumn{4}{c|}{48.4(60.3/0.6)$|$1.2}                                                & \multicolumn{4}{c|}{50.4(62.8/0.8)$|$1.6}                                               & \multicolumn{4}{c}{49.8(62.0/0.9)$|$1.8}                                               \\ \hline
\textbf{MS}      & \multicolumn{2}{c|}{1000}                  & \multicolumn{2}{c|}{2000}                 & \multicolumn{2}{c|}{1000}                 & \multicolumn{2}{c|}{2000}                 & \multicolumn{2}{c|}{1000}                 & \multicolumn{2}{c}{2000}                 \\ 
SCR              & \multicolumn{2}{c|}{39.9(39.2/42.9)$|$40.9}  & \multicolumn{2}{c|}{42.9(43.5/40.8)$|$42.1} & \multicolumn{2}{c|}{40.7(42.2/34.7)$|$38.1} & \multicolumn{2}{c|}{46.3(47.2/42.8)$|$44.9} & \multicolumn{2}{c|}{40.4(40.6/39.7)$|$40.1} & \multicolumn{2}{c}{44.5(43.8/47.3)$|$45.5} \\
SCR$_{ft}$            & \multicolumn{2}{c|}{45.4(48.9/31.2))$|$38.1} & \multicolumn{2}{c|}{49.7(53.7/33.7)$|$54.6} & \multicolumn{2}{c|}{47.4(50.6/34.8)$|$41.2} & \multicolumn{2}{c|}{49.2(53.2/33.4)$|$41.0} & \multicolumn{2}{c|}{40.2(45.2/32.8)$|$38.0} & \multicolumn{2}{c}{47.0(49.2/38.5)$|$43.5} \\
OCM              & \multicolumn{2}{c|}{43.4(43.8/42.2)$|$42.9}  & \multicolumn{2}{c|}{43.2(43.6/41.8)$|$42.7} & \multicolumn{2}{c|}{39.9(39.8/40.6)$|$40.2} & \multicolumn{2}{c|}{43.6(43.9/42.3)$|$43.1} & \multicolumn{2}{c|}{39.5(38.9/41.8)$|$40.3} & \multicolumn{2}{c}{43.5(42.9/46.0)$|$44.3} \\
OCM$_{ft}$            & \multicolumn{2}{c|}{46.8(49.8/34.8)$|$40.9}  & \multicolumn{2}{c|}{47.8(49.8/39.7)$|$44.2} & \multicolumn{2}{c|}{46.1(48.8/35.1)$|$40.8} & \multicolumn{2}{c|}{48.7(51.8/36.3)$|$42.7} & \multicolumn{2}{c|}{44.1(46.1/36.3)$|$40.6} & \multicolumn{2}{c}{46.3(48.6/37.2)$|$42.1} \\
DVC              & \multicolumn{2}{c|}{43.0(42.8/43.8)$|$43.3}  & \multicolumn{2}{c|}{45.2(45.8/42.8)$|$44.2} & \multicolumn{2}{c|}{39.4(38.6/42.5)$|$40.5} & \multicolumn{2}{c|}{40.1(39.7/41.9)$|$40.8} & \multicolumn{2}{c|}{41.6(42.6/37.4)$|$39.8} & \multicolumn{2}{c}{45.4(45.6/44.8)$|$45.2} \\
DVC$_{ft}$            & \multicolumn{2}{c|}{48.0(50.3/39.0)$|$43.9}  & \multicolumn{2}{c|}{50.3(53.7/36.9)$|$43.7} & \multicolumn{2}{c|}{41.1(41.6/38.9)$|$40.2} & \multicolumn{2}{c|}{42.4(43.9/36.4)$|$39.8} & \multicolumn{2}{c|}{45.3(48.6/31.9)$|$38.5} & \multicolumn{2}{c}{47.8(50.8/36.2)$|$42.2} \\ \hline
\rowcolor{WhiteSmoke}\textbf{Ours}    & \multicolumn{4}{c|}{\textbf{55.2}(55.6/53.7)$|$\textbf{54.6}}                                              & \multicolumn{4}{c|}{\textbf{54.3}(56.2/46.8)$|$\textbf{51.1}}                                             & \multicolumn{4}{c}{\textbf{52.2}(52.6/50.8)$|$\textbf{51.7}}                                             \\ \hline
\end{tabular}
\caption{\textbf{The quantitative analysis of dataset partition $\textbf{80\%}+\textbf{2\%}\times\textbf{10}$.} Class-wise accuracy ($\textbf{Acc}$) by end of the training in terms of all classes, base classes, and novel classes and Harmonic accuracy (\textbf{HM}) are illustrated. \textbf{MS} and $_{ft}$ mean the example memory size and finetuning versions. The best results are marked in \textbf{bold}.}
\label{80-2-10}
\end{table*}

\begin{table*}[]\centering
\scriptsize
\renewcommand\arraystretch{1.1}
\begin{tabular}{l|clcl|clclclcl}
\hline
\textbf{Methods} & \multicolumn{4}{c|}{\textbf{CORE-50}}                                                 & \multicolumn{4}{c}{\textbf{CIFAR100}}                                                 & \multicolumn{4}{c}{\textbf{Mini-ImageNet}}                                           \\ \hline
\textbf{Metrics} & \multicolumn{4}{c|}{\textbf{Acc}(base/novel)$|$\textbf{HM}}                                               & \multicolumn{4}{c|}{\textbf{Acc}(base/novel)$|$\textbf{HM}}                                               & \multicolumn{4}{c}{\textbf{Acc}(base/novel)$|$\textbf{HM}}                                               \\ \hline
ALICE            & \multicolumn{4}{c|}{39.5(46.2/29.5)$|$36.0}                                             & \multicolumn{4}{c|}{42.5(53.5/25.9)$|$34.9}                                             & \multicolumn{4}{c}{41.1(51.4/25.7)$|$34.3}                                             \\ \hline
PASS             & \multicolumn{4}{c|}{35.2(58.3/0.8)$|$1.6}                                               & \multicolumn{4}{c|}{37.9(62.6/1.0)$|$2.0}                                               & \multicolumn{4}{c}{37.2(61.2/1.1)$|$2.2}                                               \\ \hline
\textbf{MS}      & \multicolumn{2}{c|}{1000}                 & \multicolumn{2}{c|}{2000}                 & \multicolumn{2}{c|}{1000}                 & \multicolumn{2}{c|}{2000}                 & \multicolumn{2}{c|}{1000}                 & \multicolumn{2}{c}{2000}                 \\ \hdashline
SCR              & \multicolumn{2}{c|}{38.6(37.2/40.6)$|$38.8} & \multicolumn{2}{c|}{38.6(39.2/37.6)$|$38.4} & \multicolumn{2}{c|}{36.9(38.8/34.1)$|$36.3} & \multicolumn{2}{c|}{40.7(42.8/37.6)$|$40.0} & \multicolumn{2}{c|}{34.4(34.6/34.2)$|$34.4} & \multicolumn{2}{c}{35.1(38.6/29.8)$|$33.6} \\
SCR$_{ft}$           & \multicolumn{2}{c|}{36.6(42.4/27.9)$|$33.7} & \multicolumn{2}{c|}{41.3(49.8/28.6)$|$36.3} & \multicolumn{2}{c|}{38.6(49.2/22.8)$|$31.2} & \multicolumn{2}{c|}{41.1(52.8/23.7)$|$32.7} & \multicolumn{2}{c|}{37.9(43.2/30.0)$|$35.4} & \multicolumn{2}{c}{39.5(43.7/33.2)$|$37.7} \\
OCM              & \multicolumn{2}{c|}{38.3(38.8/37.6)$|$38.2} & \multicolumn{2}{c|}{41.1(42.1/39.6)$|$40.8} & \multicolumn{2}{c|}{36.3(36.5/35.9)$|$36.2} & \multicolumn{2}{c|}{41.4(42.7/39.6)$|$41.1} & \multicolumn{2}{c|}{35.9(35.2/37.1)$|$36.1} & \multicolumn{2}{c}{39.5(43.7/33.2)$|$37.7} \\
OCM$_{ft}$            & \multicolumn{2}{c|}{37.7(43.1/29.7)$|$35.2} & \multicolumn{2}{c|}{43.3(47.9/36.4)$|$41.4} & \multicolumn{2}{c|}{39.7(45.7/30.7)$|$36.7} & \multicolumn{2}{c|}{40.9(45.9/33.4)$|$38.7} & \multicolumn{2}{c|}{36.8(39.8/32.2)$|$35.4} & \multicolumn{2}{c}{41.5(43.5/38.4)$|$40.7} \\
DVC              & \multicolumn{2}{c|}{38.1(37.8/38.6)$|$38.2} & \multicolumn{2}{c|}{40.5(41.8/38.6)$|$40.1} & \multicolumn{2}{c|}{37.5(37.2/38.1)$|$37.7} & \multicolumn{2}{c|}{40.2(41.6/38.2)$|$39.8} & \multicolumn{2}{c|}{34.4(32.9/36.7)$|$34.7} & \multicolumn{2}{c}{37.1(35.4/39.6)$|$37.4} \\
DVC$_{ft}$            & \multicolumn{2}{c|}{39.8(45.9/30.7)$|$36.8} & \multicolumn{2}{c|}{41.8(46.3/35.0)$|$39.9} & \multicolumn{2}{c|}{37.8(42.1/31.4)$|$35.9} & \multicolumn{2}{c|}{40.6(44.8/34.3)$|$38.9} & \multicolumn{2}{c|}{35.1(38.9/29.4)$|$33.4} & \multicolumn{2}{c}{37.7(38.7/36.1)$|$37.4} \\ \hline
\rowcolor{WhiteSmoke}\textbf{Ours}    & \multicolumn{4}{c|}{\textbf{49.5}(49.1/50.2)\textbf{49.6}}                               & \multicolumn{4}{c|}{\textbf{47.6}(51.6/41.7)$|$\textbf{46.1}}                                    & \multicolumn{4}{c}{\textbf{49.1}(53.8/42.1)$|$\textbf{47.2}}                              \\ \hline
\end{tabular}
\caption{\textbf{The quantitative analysis of dataset partition $\textbf{60\%}+\textbf{2\%}\times\textbf{20}$.} Class-wise accuracy ($\textbf{Acc}$) by end of the training in terms of all classes, base classes, and novel classes and Harmonic accuracy (\textbf{HM}) are illustrated. \textbf{MS} and $_{ft}$ mean the example memory size and finetuning versions. The best results are marked in \textbf{bold}.}
\label{60-2-20}
\end{table*}

% \\ \hspace*{\fill} \\
\textbf{\large{Appendix C: Results and Analysis on Different the Base Session Training Strategies}}
% \\ \hspace*{\fill} \\

The stability and plasticity dilemma is a thorny problem in the area of continual learning. To deal with this dilemma, previous NE-CL \cite{ne-cl-pa, ne-cl-da} and FS-CL \cite{fs-cl-open, fs-cl-s3c} methods employ self-supervised learning \cite{ssl} and class and data augmentation to learn task-agnostic and transferable representations. For the problem of NO-CL, the base session training strategies also matter for the stability and plasticity dilemma. For fair comparisons, similar to \cite{scr, dvc, ocm}, we also employ supervised contrastive (SC) learning. Here, we provide two training strategies. Concretely, we add the extra self-supervised learning loss \cite{ssl_loss} (+SSL) like \cite{ne-cl-pa, fs-cl-s3c} and use the data augmentation strategy (+DA) proposed by \cite{ne-cl-da, fs-cl-open}. The results in Table \ref{aug} show that the elaborately designed pre-training strategies improve the accuracy both in the base and novel classes. Therefore, developing a more robust pre-training strategies is a promising way for the proposed NO-CL problem.

\begin{table}[]\centering
\small
\renewcommand\arraystretch{1.1}
\begin{tabular}{ll|cl|cl}
\hline
\multicolumn{2}{l|}{\textbf{Ablations}} & \multicolumn{2}{c|}{\textbf{CIFAR100}}    & \multicolumn{2}{c}{\textbf{Mini-ImageNet}} \\ \hline
\multicolumn{2}{l|}{\textbf{Metrics}}        & \multicolumn{2}{c|}{\textbf{Acc}(base/novel)}   & \multicolumn{2}{c}{\textbf{Acc}(base/novel)}     \\ \hline
\multicolumn{2}{l|}{Ours(+CE)}                    & \multicolumn{2}{c|}{45.8(50.0/39.6)} & \multicolumn{2}{c}{47.7(52.6/40.3)}   \\\hline
\multicolumn{2}{l|}{Ours(+SC)}                    & \multicolumn{2}{c|}{48.6(52.4/42.9)} & \multicolumn{2}{c}{50.7(56.1/42.6)}   \\
\multicolumn{2}{l|}{+SSL}               & \multicolumn{2}{c|}{50.4(53.8/45.2)} & \multicolumn{2}{c}{52.2(57.3/44.6)}   \\ 
\multicolumn{2}{l|}{+DA}               & \multicolumn{2}{c|}{51.2(55.7/44.6)} & \multicolumn{2}{c}{53.2(58.2/45.8)}   \\ \hline
\end{tabular}
\caption{The results of Different training strategies on CIFAR100 and Mini-ImageNet. SSL and DA mean self-supervised learning and data augmentation, respectively.}
\label{aug}
\end{table}

\begin{figure}[]
\centering
\includegraphics[width=0.7\columnwidth]{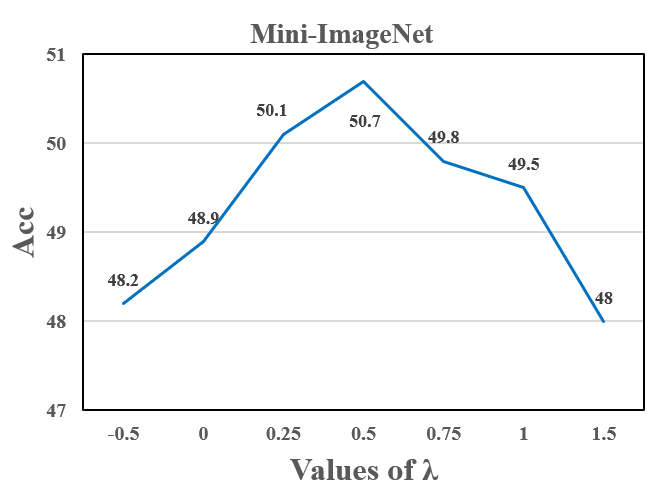}
\caption{Quantitative results of varying $\lambda$.}
\label{fig1}
\vspace{-0.8em}
\end{figure}

\begin{figure}[]
\centering
\includegraphics[width=0.4\textwidth]{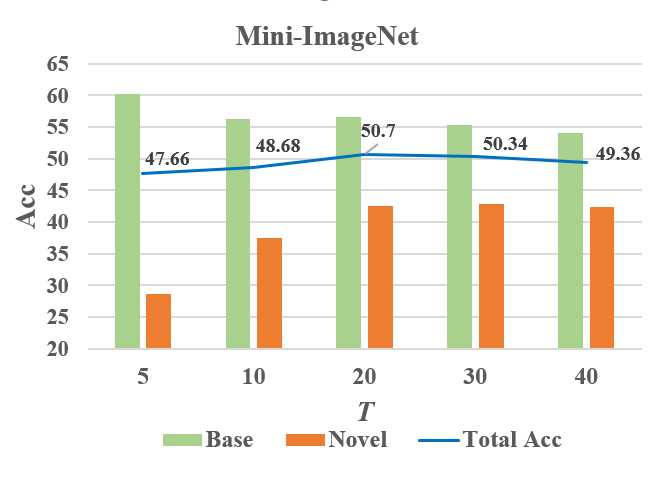}
\caption{Quantitative results of varying $T$.}
\label{fig2}
% \vspace{-0.8em}
\end{figure}
\begin{figure*}[htb]
\centering
\includegraphics[width=1.0\textwidth]{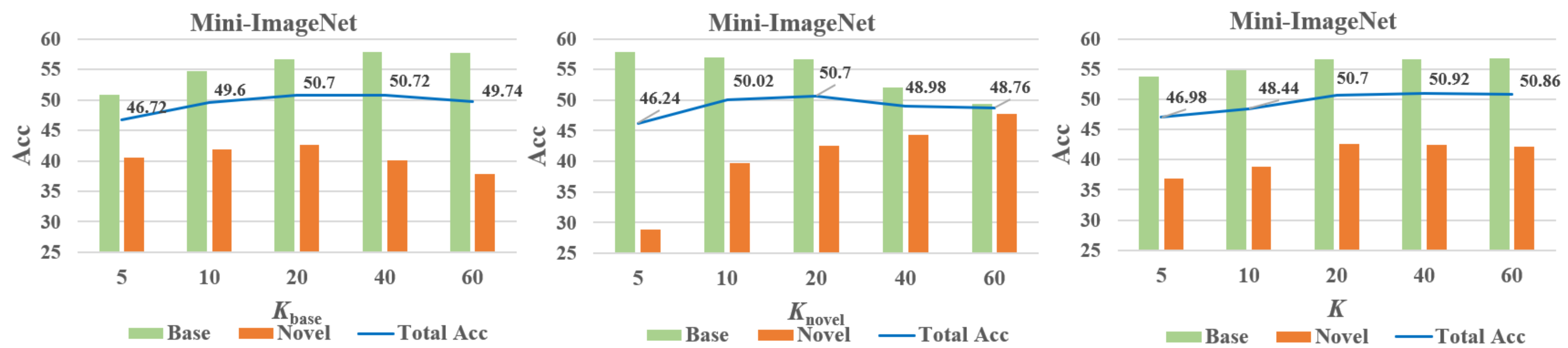}
\caption{Quantitative results of varying $K_0$, $K_1$, and $K_2$}
\label{fig3}
% \vspace{-0.8em}
\end{figure*}

% \begin{figure}[htb]
% \centering
% % \includegraphics[width=130, angle=-90]{prototype}
% \includegraphics[width=0.7\columnwidth]{N}
% \caption{Quantitative results of varying $N$}
% \label{fig4}
% % \vspace{-0.8em}
% \end{figure}

\begin{table}[]\centering
\small
\begin{tabular}{l|c|c}
\hline
                   & \textbf{CIFAR100}           & \textbf{Mini-ImageNet}     \\ \hline
\textbf{Ablations} & \textbf{Acc}(base/novel)/\textbf{HM} & \textbf{Acc}(bse/novel)/\textbf{HM} \\ \hline
w/ 256             & 44.8(49.5/37.9)/42.9        & 47.4(53.6/38.1)/44.5       \\ \cline{1-1}
w/ 1024            & 47.4(51.8/40.8)/45.7        & 49.7(55.4/41.1)/47.2       \\ \cline{1-1}
w/ 2048            & 48.6(52.4/42.9)/47.2        & 50.7(56.1/42.6)/48.4       \\ \cline{1-1}
w/ 3074            & 48.4(52.3/42.7)/47.0        & 50.8(56.3/42.6)/48.4       \\ \hline
\end{tabular}
\caption{Quantitative results of varying dimension of hyperdimensional embedding.}
\label{hp}
\end{table}
\begin{figure*}[] \centering
    \centering
        \subfigure[CORE-50]{\includegraphics[width=0.3\linewidth]{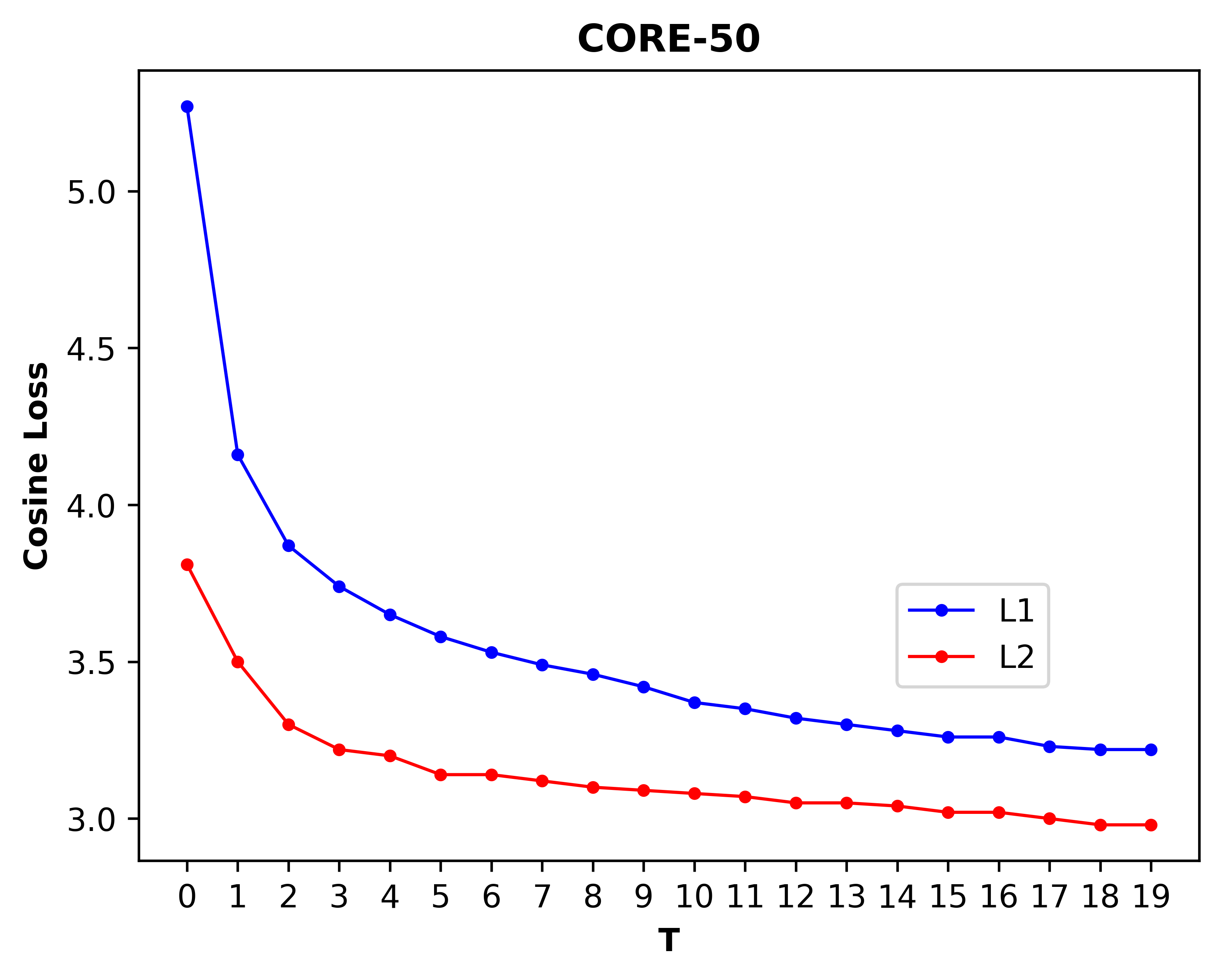}}
        \subfigure[CIFAR100]{\includegraphics[width=0.3\linewidth]{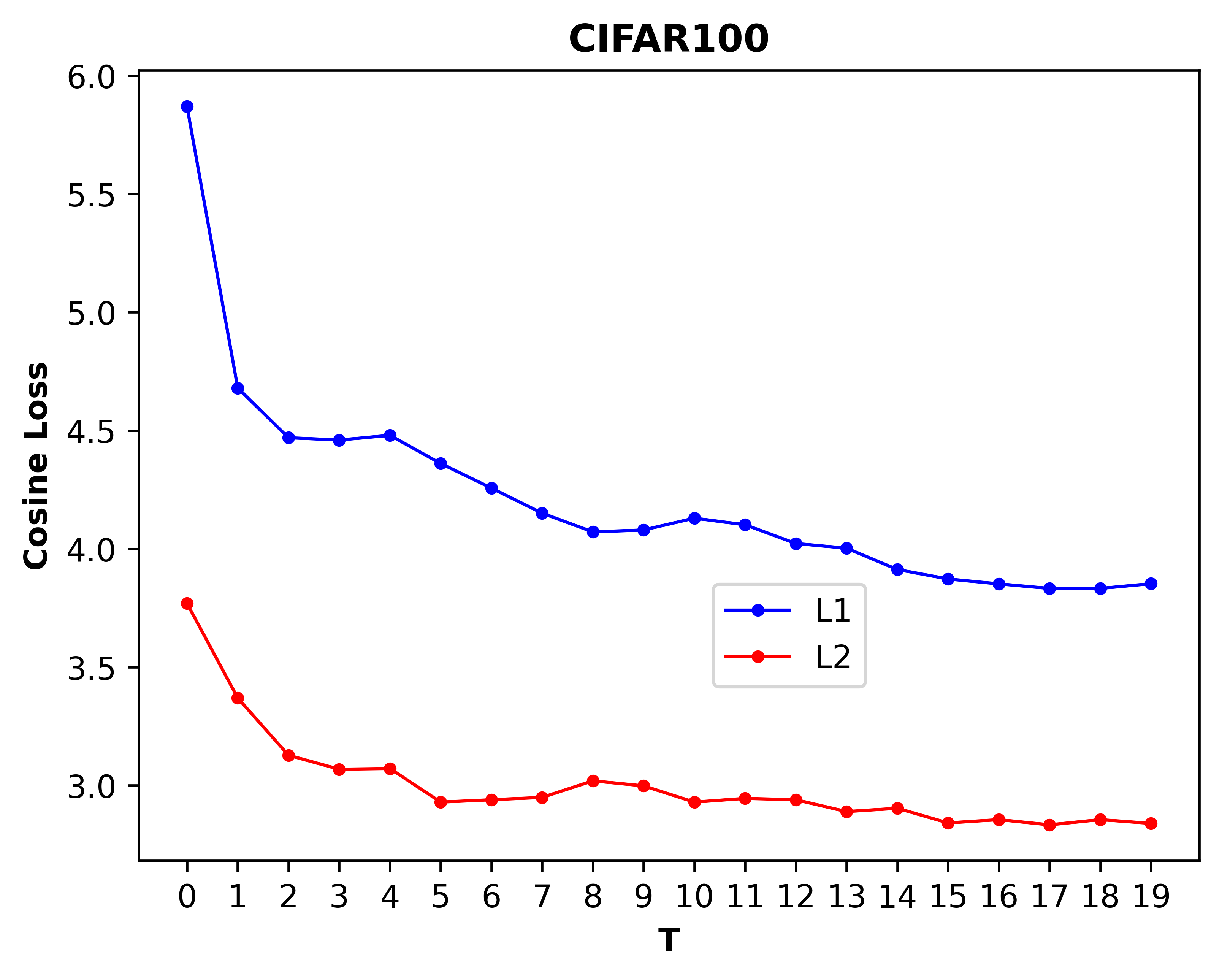}}
        \subfigure[Mini-ImageNet]{\includegraphics[width=0.3\linewidth]{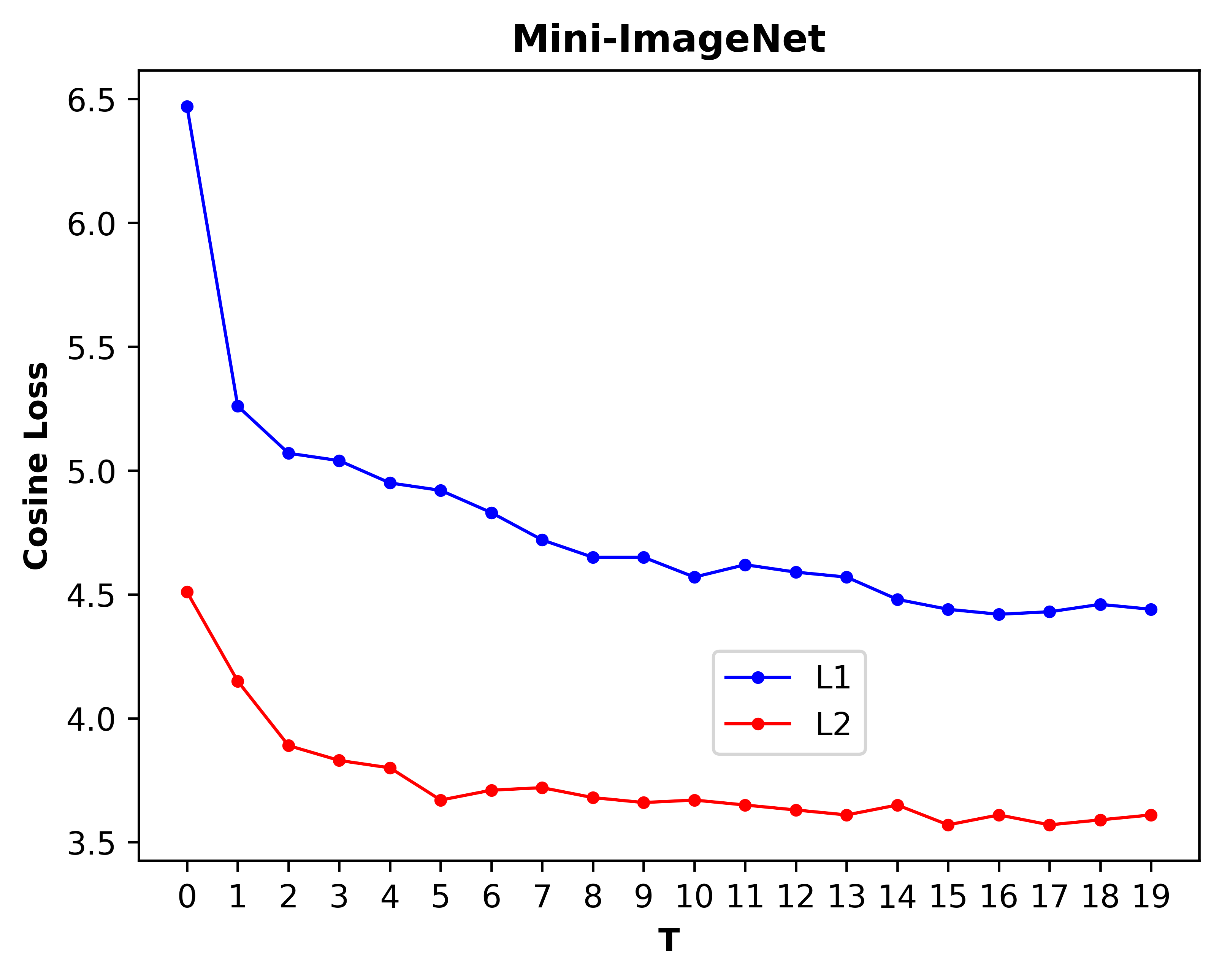}}
        \caption{Training losses of bi-level optimization procedure}%esada
\label{fig4}
\end{figure*}
\begin{table*}[]\centering
\small
\begin{tabular}{l|cccccc|cccccc}
\hline
                 & \multicolumn{6}{c|}{\textbf{CIFAR100}} & \multicolumn{6}{c}{\textbf{Mini-ImageNet}} \\ \hline
\textbf{Metrics} & ALICE & SSRE & SCR & DVC & OCM  & Ours & ALICE  & SSRE  & SCR  & DVC  & OCM  & Ours \\ \hline
\textbf{Time}(s)          & 512   & 291  & 165 & 126 & 561  & \textbf{35}   & 793    & 457   & 254  & 194  & 831  & \textbf{61}   \\ \hline
\textbf{Memory}(GB)       & 1.9   & 3.2  & 1.8 & 1.6 & 12.8 & \textbf{1.4}  & 4.4    & 6.8   & 4.1  & 2.8  & 21.4 & \textbf{1.9}  \\ \hline
\end{tabular}
\caption{Quantitative comparisons of computation overhead in terms of online training time and memory footprint.}
\label{comp}
\end{table*}

% \\ \hspace*{\fill} \\
\textbf{\large{Appendix D: Hyper Parameter Analysis}}
% \\ \hspace*{\fill} \\

Here, we analyse hyperparameters including feature transform coefficient $\lambda$, online iteration $T$, and the number of sampled prototypes $K$. We provide quantitative results on the Mini-ImageNet dataset in Figures \ref{fig1}, \ref{fig2}, and \ref{fig3}. Also, the experiments of the varying dimension of hyperdimensional embedding are conducted in Table \ref{hp}.

In Figure \ref{fig1}, we can see that $\lambda > 1$ leads to degraded performance as the feature distribution is more concentrated close to 0. Meanwhile, decreasing too much $\lambda$ makes the distribution scattered and less aligned to the calibrated Gaussian distribution. Therefore, we set $\lambda$ as 0.5 in our experiment.

In Figure \ref{fig2}, we vary online iteration $T$. Less $T$ iterations harm the network to accommodate online novel classes by refining hyperdimensional prototypes and aligning the projection module. Meanwhile, more online iterations only lead to slight degradation, which validates the plasticity of our method. Therefore, we set $T$ to 20 to achieve the stability-plasticity trade-off.

In Figure \ref{fig3}, we vary the number of sampled prototypes $K$. Concretely, we vary the number of sampled prototypes of base classes $K_{base}$, novel classes $K_{novel}$, and all classes $K$. We can see that the imbalance sampling of classes leads to performance degradation, which is similar to the class imbalance problem \cite{luc, lsil}. Also, increasing $K$ does not bring many gains while inducing computation overheads. Therefore, we set $K=20$.

From Table \ref{hp}, we can learn that increasing the dimension of hyperdimensional embedding benefits the proposed method while too large dimension brings little gain. Therefore, we set the dimension of hyperdimensional embedding as 2048 in our experiments.

% \\ \hspace*{\fill} \\
\textbf{\large{Appendix E: Computation Overhead Analysis}}
% \\ \hspace*{\fill} \\

For computation overheads during online learning, which is usually considered in OCL scenarios \cite{tl_ocl}, we provide more quantitative comparisons to OCL, NE-CL, and FS-CL methods in Table \ref{comp}. The batchsize of example-based methods is set as 10. As we only align prototypes by finetuning the projection module, which is much more efficient compared with training the whole network. Therefore, our method has clear advantages on computation overheads for online continual learning. Meanwhile, the bi-level optimization quickly converges as shown in Figure \ref{fig4}.

\bibliography{aaai24}

\end{document}